\documentclass[11pt,a4paper]{article}

\usepackage{emnlp2021}
\usepackage{times}
\usepackage{latexsym}

\usepackage{graphicx}
\usepackage{multirow, makecell}
\usepackage{listings}
\usepackage{booktabs}
\newcommand{\tabitem}{~~\llap{\textbullet}~~}

\usepackage{microtype}
\usepackage{enumitem}
\usepackage{amsmath}
\usepackage{bm}
\usepackage{caption}
\usepackage{hyperref}

\newcommand\Tstrut{\rule{0pt}{2.6ex}}       
\newcommand\Bstrut{\rule[-0.9ex]{0pt}{0pt}} 
\newcommand{\TBstrut}{\Tstrut\Bstrut} 

\title{Does Commonsense help in detecting Sarcasm?}

\author{Somnath Basu Roy Chowdhury \qquad
 Snigdha Chaturvedi \\
 \texttt{\{somnath, snigdha\}@cs.unc.edu} \\
 UNC Chapel Hill}

\date{}

\begin{document}
\maketitle
\begin{abstract}
 Sarcasm detection is important for several NLP tasks such as sentiment identification in product reviews, user feedback, and online forums. It is a challenging task requiring a deep understanding of language, context, and world knowledge. In this paper, we investigate whether incorporating commonsense knowledge helps in sarcasm detection.  For this, we incorporate commonsense knowledge into the prediction process using a graph convolution network with pre-trained language model embeddings as input. Our experiments with three sarcasm detection datasets indicate that the approach does not outperform the baseline model. We perform an exhaustive set of experiments to analyze where commonsense support adds value and where it hurts classification. Our implementation is publicly available at: \href{https://github.com/brcsomnath/commonsense-sarcasm}{https://github.com/brcsomnath/commonsense-sarcasm}.
\end{abstract}

\section{Introduction \& Related Work}

The topic of sarcasm has received attention in various research fields like linguistics~\cite{utsumi2000verbal},  psychology~\cite{gibbs1986psycholinguistics, kreuz1989sarcastic} and the cognitive sciences~\cite{gibbs2007irony}. Identifying sarcasm is essential to understanding the opinion and intent of a user in downstream tasks like opinion mining, sentiment classification, etc.  Initial approaches for this task~\cite{kreuz1989sarcastic} mostly relied on handcrafted features to capture the lexical and contextual information. On similar lines, the efficacy of special characters, emojis and n-gram features in the discrimination task have also been studied~\cite{carvalho2009clues, lukin2017really}. 

\begin{figure}[t!]
	\includegraphics[width=0.5\textwidth]{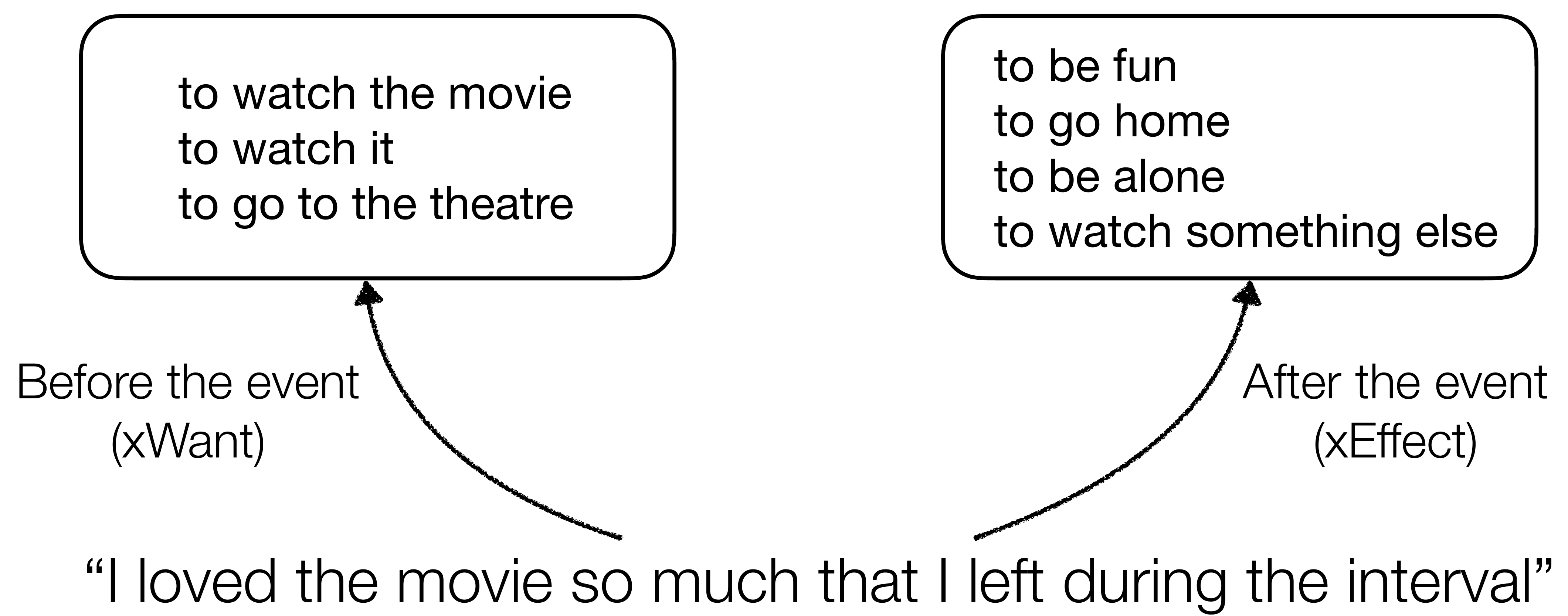}
	\centering
	\caption{COMET output for the sentence ``I loved the movie so much that I left during the interval". The commonsense sequences capture the contrast between intent and action of the subject.}
	\label{fig:1}
\end{figure}

In recent years, this task has gained traction in the machine learning and computational linguistic community~\cite{davidov2010semi,  gonzalez2011identifying, riloff2013sarcasm, maynard2014cares, wallace2014humans, ghosh2015sarcastic, joshi2015harnessing, muresan2016identification, amir2016modelling, mishra2017harnessing, ghosh2017magnets, chakrabarty2020r}. Several approaches have studied the role of context in this sarcasm detection task~\cite{ghosh2020report}. However, none of the previous works have explored the idea of incorporating commonsense knowledge in sarcasm detection.  Common sense has been used in several natural-language based tasks like controllable story generation~\cite{zhang2020story, brahman2020modeling}, sentence classification~\cite{chen2019deep}, question answering~\cite{dzendzik2020q}, natural language inference~\cite{annervaz2018learning, wang2019improving} and other related tasks but not for sarcasm detection. We hypothesize that commonsense knowledge, capturing general beliefs and world knowledge, can prove instrumental in understanding sarcasm. For example in Figure~\ref{fig:1}, for the event ``I loved the movie so much that I left during the interval" (an example of sarcasm with polarity contrast), we show how commonsense is able to capture the contrast between the intentions of the subject before and during the event. Incorporating such commonsense knowledge ideally should make it easier for the learning model to detect sarcasm where it is not apparent from the input.  

With this motivation, we study the utility of common sense information for sarcasm detection. For this, we leverage COMET~\cite{bosselut2019comet} to extract the relevant {social} commonsense information for a sentence.
Given an event, COMET provides likely scenarios relating to various attributes like intent of the subject, effect on the object etc.

We use a GCN~\cite{kipf2016semi} based model for infusing commonsense knowledge in the sarcasm detection task. Our experiments reveal that the commonsense augmented model performs at par with the baseline model. We perform an array of analysis experiments to identify where the commonsense infused model outperforms the baseline and where it fails.

\begin{figure}[t!]
	\includegraphics[width=0.45\textwidth]{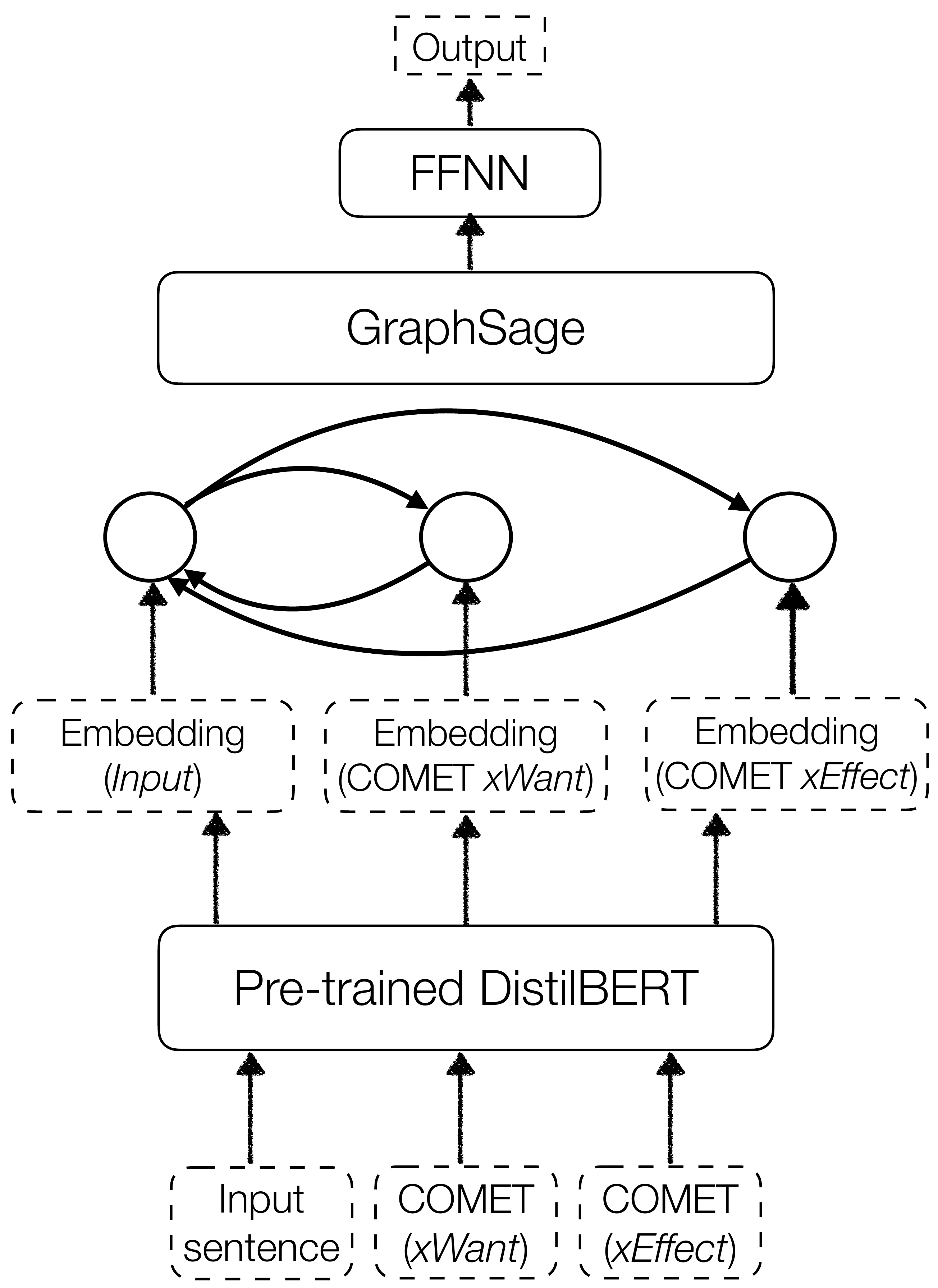}
	\centering
	\caption{Proposed model architecture. Representations of the input sentence along with two COMET sequences are retrieved from pre-trained DistilBERT that are used to initialize a GCN. Post training, node representations of the graph is passed through a fully-connected neural network to generate the output.}
	\label{fig:6}
\end{figure}

\section{Model}
We use a graph convolution-based model to enable incorporation of COMET sequences given an input sentence. The sentence representations are retrieved from the pre-trained encoder of our baseline model. Our \textbf{baseline model} consists of a Transformer~\cite{vaswani2017attention} based  DistilBERT~\cite{sanh2019distilbert} encoder followed by a feed-forward neural network (FFNN). DistilBERT is a light-weight encoder, which enables faster training, while achieving similar performance as other Transformer based encoders.

The model is shown in Figure~\ref{fig:6}. For every input instance, a graph is formed with edges between the input sentence and COMET sequences. No edges are present between individual COMET sequences. Sentence embeddings retrieved from the baseline DistilBERT form the initial graph embeddings. The intuition behind leveraging a graph-based architecture was to enable information flow between the representations of the input sentence and COMET sequences, thereby reducing the domain discrepancy between them.

The graph is then fed into a GraphSage~\cite{hamilton2017inductive} network which produces the node embedding vector $V \in {\rm I\!R}^{(M + 1) \times N}$, where $M$ is the number of COMET sequences and $N$ is the output dimension of the GCN. The node embedding vector $V$ is then forwarded to a fully connected neural network layer to produce the final output. In section~\ref{gcn-exp}, we experiment with different edge configurations and observe how each edge configuration affects the downstream performance.


We experimented with another model that incorporated COMET sequences with an attention-mechanism. In that model, the representation of the input sentence was concatenated with an aggregate representation of the COMET sequences, formed in an attentive fashion. 
Its performance was not better than the GCN-based model, so we do not describe it here.

\begin{table}[t!]
	\begin{center}
	
		\resizebox{0.45\textwidth}{!}{
		\begin{tabular}{l|c|c}
			\toprule[1pt]
			\textsc{Dataset} & \textsc{Training} & \textsc{Test} \TBstrut\\
			\midrule[1pt]
			{SemEval Irony } & 3833 & 958 \Tstrut\\
			{News Headline} & 27691 & 27691 \\

			{FigLang 2020 Reddit} & 3520 & 880 \\
			\bottomrule[1pt]
		\end{tabular}
		}
		\caption{Number of training samples in train/test split for each dataset.}
		\label{table0}
	\end{center}
\end{table}

\begin{table*}[t!]
	\centering
			\begin{tabular}{p{0.34\textwidth}|c|c|c} 
				\toprule[1pt]
				\multirow{2}{*}{\textsc{Approach}}
				& \multirowcell{2}{News \\ Headlines}
				& \multirowcell{2}{SemEval\\ Irony}
				& \multirowcell{2}{FigLang 2020\\ (Reddit)}\Tstrut\\
				& & & \Bstrut\\
				\midrule[1pt]
				Baseline & 96.13\% & 69.09\%& 67.95\%\Tstrut\\
				GCN (\textit{bidirectional} edges) & 96.16\% & 67.88\%&  67.50\%\\
				GCN (input $\rightarrow$ COMET edges) & 96.14\% & 68.66\%&  67.35\%\\
				GCN (COMET $\rightarrow$ input edges) & 96.18\% & 68.40\%&  67.54\%\\
				\bottomrule[1pt]
		\end{tabular}
		\caption{Accuracy of the baseline DistilBERT and GCN model (in various edge configurations). We do not observe any significant change in sarcasm detection performance with the incorporation of commonsense sequences.}
		\label{table5}
\end{table*}
\section{Experimental Setup}
\label{sec:exp}

We evaluate  the  models on three datasets (a) \textbf{Irony detection SemEval task}: \citeauthor{van2018semeval} (\citeyear{van2018semeval}) conducted a SemEval task for irony detection considering an utterance in isolation. They also released a secondary task where the sarcastic samples were classified into \textit{three} broad categories: verbal irony with polarity contrast, situational irony, and others. (b) \textbf{News Headlines dataset} \cite{misra2019sarcasm}: contains sarcastic news headlines from \textit{TheOnion} 
and non-sarcastic ones from \textit{HuffPost}. (c)  \textbf{FigLang 2020 Sarcasm detection task}: We experiment on the Reddit dataset of the shared task introduced by \citeauthor{ghosh2020report} (\citeyear{ghosh2020report}). The statistics of the datasets are specified in Table~\ref{table0}.

All the aforementioned datasets are balanced. We report our results by randomly splitting into training and test set, and averaging the accuracy over 5 iterations.  In our experiments, we incorporate a subset of COMET predicates (\textit{xWant} and \textit{xEffect}) related to the subject in a sentence.

  \begin{table}[t!]
	\begin{center}
		\resizebox{0.42\textwidth}{!}{
		\begin{tabular}{l|c} 
			\toprule[1pt]
			{Edge configuration} & {Performance} \TBstrut\\
			\midrule[1pt]
			GCN (\textit{bidirectional}) & 67.27\%\Tstrut\\
			GCN (COMET $\rightarrow$ input)  & 55.00\% \\
			GCN (input $\rightarrow$ COMET)  & 67.36\% \\
			\bottomrule[1pt]
		\end{tabular}
		}
		\caption{Performance of the proposed model for different edge configurations. We observe a sharp performance drop in (COMET $\rightarrow$ input) configuration.}
		\label{table10}
	\end{center}
\end{table}

\section{Results}
\label{gcn-exp}

We report the classification accuracy of the models for all datasets in Table~\ref{table5}. The baseline denotes the DistilBERT performance. We see a high performance in the \textit{News headline} dataset where the sentences are self-contained and language is not noisy. We see relatively lower performance of the baseline in FigLang 2020 Reddit, where we ignored the available context. Performance in SemEval dataset is low due to noisy tweets. 

We conduct an ablation study with three configurations of the graph edges (a) \textit{bidirectional} edges (b) edges from \textit{input} $\rightarrow$ \textit{COMET} sequences and (c) edges from \textit{COMET sequences }$\rightarrow$ \textit{input}. The results of the GCN-based model in different settings are shown in Table~\ref{table5}. The performance of the GCN model is at par with the baseline and varying edge configurations doesn't have any effect on the downstream performance.  
COMET produces sequences for an input event in the following format.

\noindent\textit{Input}: ``they should have put \$125 million termination payout in each of their contract"\\
\textit{xWant} : \textit{to save money}

\noindent We wish to have more complete COMET sentences like: \textit{they wanted to save money}.
In order to improve the setup, we replace COMET sequences with complete sentences  
of the form:
\texttt{[subject] [MASK] [raw COMET sequence]}.
We replace the \texttt{[subject]} placeholder with the  SUBJECT POS tag in the input sentence. We leverage a pre-trained
BERT model to predict the unknown \texttt{[MASK]} word. All reported results use this setup.

  We examine whether the COMET representations leverage information from the input in the GCN setup by \textbf{removing the input sentence representation} before the FFNN module (shown in Figure~\ref{fig:6}) and experimenting with different edge-configurations. In Table~\ref{table10}, we observe a significant performance dip with COMET$\rightarrow$input setup. This illustrates that the information flowing from input sentence to COMET sequences is more relevant.
  
\begin{table}[t!]
	\begin{center}
		\resizebox{0.35\textwidth}{!}{
		\begin{tabular}{l|c} 
			\toprule[1pt]
			{Dataset} & {Overlap} \TBstrut\\
			\midrule[1pt]
			News Headline & 99.5\%\Tstrut\\
			SemEval Irony  & 91.6\% \\
			FigLang 2020 (Reddit) & 92.7\% \\
			\bottomrule[1pt]
		\end{tabular}
		}
		\caption{Test set overlap where the output label from the GCN and DistilBERT model is the same. }
		\label{table8}
	\end{center}
\end{table}

\begin{figure*}[h!]	
    \centering
	\includegraphics[width=\textwidth, keepaspectratio]{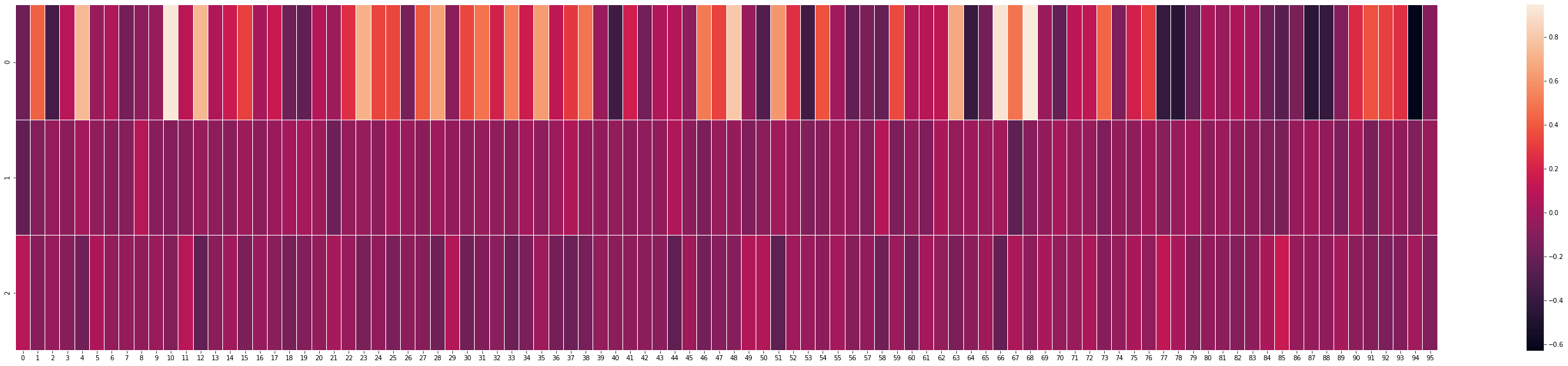}
	\caption{Visualization of gradient-based saliency tests. Darker shade denotes lower absolute values. The first row shows the features corresponding to the input sentence, and the other two rows are features from COMET sequences \textit{xWant} and \textit{xEffect}. We observe that features from input sentence (first row) receive high saliency values.}
	\label{fig:8}
\end{figure*}

  We also measure the share of instances in the test set having the same predicted label from the baseline and the model. We observe a significant overlap ($>$90\%) between the predictions of the baseline and the proposed model across all datasets in Table~\ref{table8}, illustrating that the model isn't able to tackle new instances.

  \begin{table}[t!]
 	\begin{center}
 		\begin{tabular}{l|c} 
 			\hline
 			{Occluded Element} & $\Delta$ \TBstrut\\
 			\hline
 			Input sentence & 27.99\%\Tstrut\\
 			COMET sequences  & 1.38\% \\
 			\hline
 		\end{tabular}
 		\caption{Confidence change when different segments of the input are occluded. $\Delta$ denotes the change in confidence when different parts of the input is occluded.  }
 		\label{table9}
 	\end{center}\vspace{-5pt}

 \end{table}
 \begin{figure}[t!]
	\includegraphics[width=0.5\textwidth, keepaspectratio]{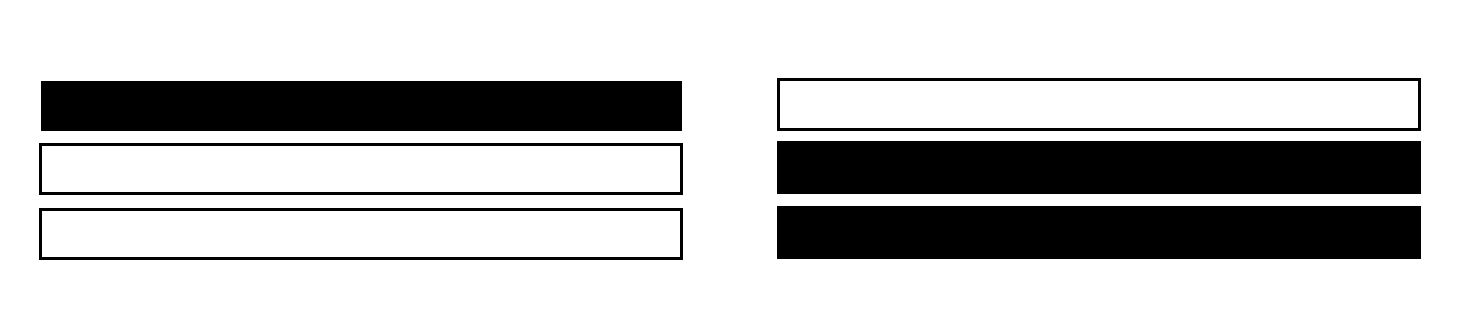}
	\centering
	\caption{Occlusion setup. First setup shows that the input sentence representation (first row) is occluded. Second setup commonsense sequence representations are occluded (second and third rows).}
	\label{fig:9}
\end{figure}

\section{Saliency Test}
We perform saliency tests to investigate whether the model is reliant on commonsense sequences while taking decisions. 

\noindent(a) \textbf{Gradient-based saliency} \cite{bastings2020elephant} measure for a feature $x_i$ given an output class $c$ is computed as $\nabla_{x_i} \mathcal{L}(y, f(x)) \cdot  x_i $, where $\mathcal{L}(\cdot, \cdot)$ is the loss function. The saliency map is shown in Figure~\ref{fig:8}. The saliency map vector has a dimension of  $3 \times 768$, where the first row showcases the saliency values of the features corresponding to the input sentence while the remaining two rows correspond to the saliency of COMET sequences. For better visualization, all values are normalized between 0-1 and average pooling is performed on adjacent blocks of 8 to form a vector of dimension $3 \times 192$. From Figure~\ref{fig:8}, it is evident that the model learns to identify important input features  but assigns similar saliency values to all COMET features.

\noindent(b) \textbf{Occlusion-based saliency} test involves occluding a part of the input and observing the change in the output probability vector. We occlude the input representation and COMET representations respectively as shown in Figure~\ref{fig:9}. The occlusion metric \cite{bastings2020elephant} is defined as $ \mathrm{E}_{x \sim \mathcal{D}} [|f_c(\bm{\mathrm{x}}) - f_c(\bm{\mathrm{x}}| x_i = 0)|]$. Table~\ref{table9} reports the results of this test. We observe that occluding the input sentence leads to a significant change in the output confidence while occluding the COMET sequences has little impact.

 These tests demonstrate that the model is more reliant on the input sentence and less on the COMET sequences for making the prediction.

\section{Efficacy of Commonsense}
In this section, we try to uncover why COMET sequences don't help in the sarcasm detection task. 
In order to identify instances where commonsense incorporation hurts the performance, we focus on samples where the model's prediction is wrong but the baseline is correct. Among these samples, we measure how many were non-sarcastic by defining a new measure \textbf{non-sarcastic class coverage},
$$\mathcal{C}^{NS}_{\textsc{GCN}} = \frac{|\{x | x \in \mathcal{S}^B_{\textsc{GCN}}, l(x) = \mathcal{NS}\}|}{|\mathcal{S}^B_{\textsc{GCN}}|}$$

	\begin{table}[t!]
		\begin{center}
			\begin{tabular}{l|c} 
				\toprule[1pt]
				\textsc{Dataset} & \text{$\mathcal{C}^{NS}_{\textsc{GCN}}$}  \TBstrut\\

				\midrule[1pt]
				
				News Headline  & 83.8\%\Tstrut\\ 
				SemEval Irony  & 64.6\% \\ 
				FigLang 2020 Reddit & 96.8\% \Bstrut\\ 
				\bottomrule[1pt]
			\end{tabular}
			\caption{$\mathcal{C}^{NS}_{\textsc{GCN}}$ statistic across different datasets.}
			\label{table11}
		\end{center}
		\vspace{-5pt}

	\end{table}
	
	\begin{table*}[t!]
    \small
	\centering
		\begin{tabular}{p{0.12\textwidth}|p{0.2\textwidth}|p{0.27\textwidth}|p{0.32\textwidth}} 
			\toprule[1pt]
			\multirowcell{2}{\textbf{Ground} \textbf{Truth}} 
			& \multirowcell{4}{\textbf{Input Sentence}} 
			& \multirowcell{4}{\textbf{COMET Support} \\[0.25em]\textit{(xWant and xEffect)}} 
			& \multirowcell{4}{\textbf{Explanations}}\\
			& &  &  \\[0.25em]
			\midrule[1pt]
			\multirowcell{2}{Non-sarcastic} 
			& 
			{\textit{@usertag} i wonder if they have that in an audio book} 
			& 
			\tabitem He gets to learns something new 
			
			\tabitem He wants to be entertained
			& 
			{COMET sequences don't add value for classifying the non-sarcastic sample.} \TBstrut\\		
			
			\hline  & & &\\[-0.75em]
			
			\multirowcell{3}{Sarcastic} 
			&
			{Going to watch a movie about murder. merry christmas ;)} 
			& 
			 \tabitem The person wants to have fun
			 
			 \tabitem The person gets tired as a result
			 &
			 {COMET sequences fail to explain the satire.} \TBstrut\\		

			\hline & & &\\[-0.75em]
			
			\multirowcell{2}{Sarcastic} 
			&
			\multirowcell{2}{final at 7am, I'm ready}
			& 
			\tabitem The person wants to go to bed 
			
			\tabitem The person has to go to work
			&
			COMET captures the contrast between the intentions and results. \TBstrut\\		
			
			\hline & & &\\[-0.75em]
			
			\multirowcell{3}{Sarcastic} 
			& 
			As a girl my reason not to put on makeup is I'm satisfied with my face
			& 
			\tabitem She wanted to look pretty 
			
			\tabitem As a result she got complements
			& 
			COMET doesn't provide relevant commonsense  for capturing polarity contrast. \TBstrut\\	
			\bottomrule[1pt]
		\end{tabular}
		\caption{Example input instances along with their ground truth label and corresponding commonsense sentences retrieved from COMET. We analyze the utility of COMET sequences described as explanations. }
		\label{examples}
\end{table*}
 \noindent where $\mathcal{S}^B_{\textsc{GCN}}$ is the set of samples in which the model predicted incorrectly while the baseline was correct, $l(\cdot)$ is the oracle function which returns the true label of an input instance $x$, and $\mathcal{NS}$ denotes the non-sarcastic class label. Results in Table~\ref{table11} demonstrate a high value of $\mathcal{C}^{NS}_{\textsc{GCN}}$ across all datasets, this indicates that the large fraction of the instances where the model was incorrect but the baseline was correct were non-sarcastic. After surveying non-sarcastic instances we infer that commonsense knowledge fails to explain non-sarcastic samples and is present as irrelevant context hurting downstream performance~\cite{petroni2020context}.

There are cases where the prediction failed either due to noisy input (prevalent in the Twitter based SemEval dataset) or subtle play of words which COMET sequences fail to explain. 

 In order to investigate the utility of commonsense for specific type of sarcasm, we form a subset of the SemEval Irony dataset with  samples only from irony with polarity contrast and non-sarcastic class by leveraging labels from the secondary SemEval task (mentioned in Section~\ref{sec:exp}). 
	$\mathcal{C}^{NS}_{\textsc{GCN}}$ for the new dataset is 57.1\%, a significant reduction from the 64.6\% in SemEval dataset in Table~\ref{table11}. We infer that commonsense is only useful in detecting sarcasm with polarity contrast but struggles with other types of sarcasm.

\section{Qualitative Analysis}
\label{appendix:relevancecs}

In this section, we analyze a few examples shown in Table~\ref{examples} and observe whether the COMET sequences are helpful in detecting sarcasm.
We have anonymized any twitter handle with ``@\textit{usertag}" to prevent any leak of private information.

\begin{itemize}[leftmargin=*]
    \topsep0mm
	\item 	In the first example of Table~\ref{examples}, the input sentence is non-sarcastic. Retrieved commonsense sequences don't capture any information that may help in prediction.
	\item 	In several instances, a sentence is sarcastic due to a subtle play of words or use of language. The commonsense based model struggles in such scenarios as COMET sequences cannot explain such events as shown in the second instance of Table~\ref{examples}. 
	\item In the third example of Table~\ref{examples},  we show that COMET sequences are able to perfectly capture the contrast between the intention and effect on the person. 
	\item  In rare cases like the fourth instance of Table~\ref{examples}, which is an example of irony with polarity contrast. It is still difficult for the commonsense model to explain the satire.
\end{itemize}

\section{Conclusion and Future work}
In this paper, we proposed the idea of integrating commonsense knowledge in the task of sarcasm detection. We observe that COMET infused model performs at par with the baseline. Through saliency tests, we observe that the model is less reliant on the commonsense representations in many cases. From our analysis, we infer that commonsense is most effective in identifying sarcasm with polarity contrast but fails to explain non-sarcastic samples or other types of sarcasm effectively, which hurts the overall performance.  In the future, we will explore the utility of other forms of external knowledge such as factual world knowledge for sarcasm detection.  We will also try to leverage commonsense to \textit{explain} why a certain remark is sarcastic. 
\bibliography{anthology,acl2021}
\bibliographystyle{acl_natbib}

\end{document}